\documentclass[conference, letterpaper, 10 pt]{ieeeconf}  % Comment this line out
\IEEEoverridecommandlockouts                              % This command is only
\usepackage[utf8]{inputenc}
\usepackage[T1]{fontenc}
\usepackage{graphicx}
\usepackage{algorithm}
\usepackage[noend]{algpseudocode}
\usepackage{mdwtab}
\usepackage{eqparbox}
\usepackage{url}
\usepackage[colorlinks=true, allcolors=blue]{hyperref}
\usepackage[usenames,dvipsnames,svgnames]{xcolor}
\usepackage{mathtools}
\usepackage{array}
\usepackage{booktabs}
\usepackage{siunitx}
\usepackage{textgreek}
\usepackage{mathptmx}
\usepackage{comment}
\usepackage{leftidx}
\usepackage{subfigure}

\newcommand{\norm}[1]{\left\lVert#1\right\rVert}

\usepackage{amsmath, amssymb, amsthm}

\title{\LARGE \bf
Domain Adaptation of Visual Policies with a Single Demonstration}

\author{Weiyao Wang$^{1}$ and Gregory D. Hager$^{1}$
\thanks{This work was supported by the U.S. National Science Foundation grants IIS-1900952 to Johns Hopkins University.}
\thanks{$^{1}$W. Wang and Gregory D. Hager are with the Department of Computer Science,
        Johns Hopkins University, Baltimore, MD 21218, USA
        {\tt\small wwang121 @ jhu.edu and hager @ cs.jhu.edu}
        }
}

\begin{document}
\maketitle

\begin{abstract}

Deploying machine learning algorithms for robot tasks in real-world applications presents a core challenge: overcoming the domain gap between the training and the deployment environment. This is particularly difficult for visuomotor policies that utilize high-dimensional images as input, particularly when those images are generated via simulation. A common method to tackle this issue is through domain randomization, which aims to broaden the span of the training distribution to cover the test-time distribution. However, this approach is only effective when the domain randomization encompasses the actual shifts in the test-time distribution. We take a different approach, where we make use of a single demonstration (a prompt) to learn policy that adapts to the testing target environment. Our proposed framework, PromptAdapt, leverages the Transformer architecture's capacity to model sequential data to learn demonstration-conditioned visual policies, allowing for in-context adaptation to a target domain that is distinct from training. Our experiments in both simulation and real-world settings show that PromptAdapt is a strong domain-adapting policy that outperforms baseline methods by a large margin under a range of domain shifts, including variations in lighting, color, texture, and camera pose. Videos and more information can be viewed at project webpage: \href{https://sites.google.com/view/promptadapt}{https://sites.google.com/view/promptadapt}.

\end{abstract}  
\section{Introduction}

Reinforcement learning (RL) has demonstrated impressive achievements in complex robotic tasks that use image observations~\cite{kalashnikov2018qt,kulhanek2019vision,hansen2021stabilizing}. At the same time, advances in simulation~\cite{todorov2012mujoco,makoviychuk2021isaac} have enabled RL algorithms to overcome their high sample complexity by training in a simulated environment before deployment on a physical robot. However, the discrepancies between the simulated data and the real-world, the well-known sim-to-real gap~\cite{zhao2020sim}, often leads to poor real-world performance~\cite{cobbe2020leveraging}. This is an extreme case of the more general problem of ensuring RL algorithms are able to accommodate domain shifts on their inputs.
The present work mainly focuses on robustness under changes in visual appearances, such as changes in lighting and texture, a significant category of environmental shifts within the sim-to-real gap.

A common approach to addressing distribution shift is to attempt to learn a visual policy that is \textit{robust} to appearance changes through domain randomization~\cite{ganin2016domain, tzeng2017adversarial, tobin2017domain, chen2021understanding}. Applying a large amount of randomization generally increases (but does not guarantee) the likelihood that visual policy will work in a novel test environment, but can also lead to increased training difficulty and decreased average policy performance. A variety of other approaches have been proposed to mitigate the difficulty of policy learning \cite{fan2021secant,ross2011reduction, ho2016generative,rusu2015policy, parisotto2015actor, chen2022system,lai2022sim}. However accommodating large domain shifts between the training environment and the deployment environment, which may be unknown during training, remains a challenge.

Instead of attempting to learn a visual policy that is robust to all unknown environmental changes, we instead pose the following question: \textit{Can a single demonstration serve as a prompt for visual domain adaptation in visuomotor policies?} Our approach to answering this question builds on a related set of ideas that use state-action trajectories, in the form of a demonstration, as a form of context or "prompt" to enable rapid adaptation to new tasks~\cite{dasari2021transformers,mandi2022towards,xu2022prompting}. In the case of domain shift, we formulate the problem as a demonstration-conditioned policy distillation challenge. The demonstration trajectory provides crucial information about both the optimal policy and the transition dynamics within the target domain. In many real-world scenarios, a human operator can offer this brief demonstration to illustrate how to complete the task under the conditions of the target domain.   

In brief, in our prompt-based domain adaptation framework, termed PromptAdapt, we first train a teacher policy with a ground truth state vector as input. We then train a Transformer-based student visual policy to imitate the teacher policy's behavior, conditioned on a demonstration with image observations obtained under various domain randomization perturbations (the prompt). During testing in a new environment, the algorithm is given a corresponding demonstration acquired in the new environment. The agent leverages the observations and policy information in the prompt to adapt to the domain shift and act accordingly. This framework is attractive because the visual policy is domain adapted with only one short trajectory segment and no explicit finetuning. It combines policy distillation and the benefits of the Transformer architecture to efficiently adapt to the target domain. 

To validate the performance of our method, we conducted tests in both simulated and real-world environments. Our results indicate that PromptAdapt is highly effective in adapting to visual domain shifts, such as changes in lighting, color, texture, and camera pose. Through extensive experiments, we demonstrate its superior performance over a range of baseline approaches. Our method consistently achieves higher cumulative rewards or success rates for both in-distribution and out-of-distribution shifts from the training set, highlighting its practicality for adapting to unknown environmental changes that may occur in real-world deployments.

\begin{figure*}[t]
\centering
\includegraphics[width=5in]{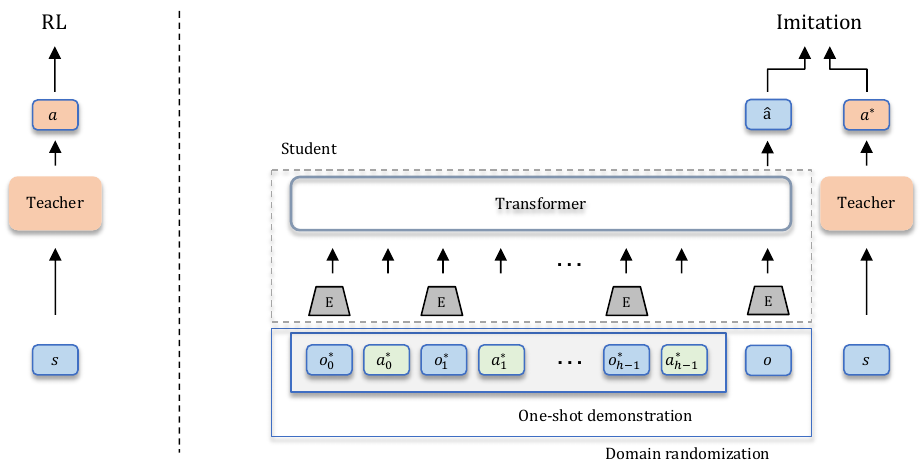}

\caption{\textbf{Left:} We first train a teacher policy using privileged ground truth state information. \textbf{Right:} We then distill the learned policy to a visual input-only student policy through imitation learning. For the student policy, our PromptAdapt architecture leverages a Transformer network to condition on a single demonstration, efficiently adapting to the target domain during testing. Same domain randomization function is applied to images in demonstration and per step observation to enable demonstration conditioned adaptation to visual appearance changes.}
\label{architecture}
\end{figure*}

\section{Related Work}

\textbf{Visual Reinforcement Learning:}
Visual reinforcement learning has gained significant attention due to its potential applicability in a wide range of real-world tasks. Methods that combine convolutional neural networks (ConvNets) with reinforcement learning algorithms have achieved remarkable success in learning policies directly from raw pixel inputs. More recent work has explored the use of unsupervised or self-supervised auxiliary tasks~\cite{finn2016deep,lin2019adaptive, lyle2021effect, he2022reinforcement} and data augmentations~\cite{hansen2021stabilizing,kostrikov2020image,hansen2021generalization,ma2022comprehensive,bertoin2022look} to learn robust and performing visual polices.

\textbf{Imitation Learning and Policy Distillation:}
Imitation learning \cite{ross2011reduction, ho2016generative} is an effective approach to transfer expert knowledge to a student policy. Policy distillation \cite{rusu2015policy, parisotto2015actor, chen2022system,lai2022sim} is a related technique where a student policy is trained to imitate a teacher policy, often with the goal of compressing or refining the teacher policy. In the context of image based RL, ~\cite{fan2021secant} proposes to disentangle between policy learning and visual generalization by policy distillation. In some approaches~\cite{chen2022system,lee2020learning, miki2022learning}, teacher can access to additional privileged information (e.g., grounded truth state) which student policy does not have. Building upon this framework, our method extends it to adapt visual policies using a single demonstrations.

\textbf{Domain Transfer for Control Policy:}
Domain transfer has been extensively studied for control policies, often in the context of sim-to-real transfer. In such scenarios, the goal is to deploy a policy trained in a simulator onto a real physical robot, where there may be discrepancies in physics, configurations, and appearances. One widely used approach to bridge this gap is domain randomization. This strategy aims to reduce the discrepancies between the source (simulated) and target (real) domains~\cite{ganin2016domain, tzeng2017adversarial}. Techniques such as~\cite{tobin2017domain, chen2021understanding} have utilized domain randomization to enhance the generalization capabilities of learned policies. They introduce variability in the simulation during training to mitigate shifts in factors such as system dynamics~\cite{lee2020learning,lim2021planar} and visual appearances~\cite{lim2022real2sim2real}. In addition to mere augmentation of the training environment, domain adaptation methods~\cite{yu2017preparing, kumar2021rma} are explored to refine usually state-based policies using test-time trajectory sequences. In contexts closely related to ours, focusing on adapting visual policies, \cite{hansen2020self} employs inverse dynamics modeling with a self-supervised learning objective to fine-tune the image encoder during test time, thereby enhancing performance. Some recent studies have delved into the use of generative adversarial training to aid sim-to-real transfer~\cite{rao2020rl, ho2020retinagan, yoneda2021invariance}.

\textbf{Meta-Learning:}
Meta-learning aims to quickly adapt to new machine learning tasks with limited data at test time~\cite{rakelly2019efficient, hospedales2021meta, wang2020generalizing, finn2017model}. In the context of reinforcement learning, meta-learning has been used to enable fast adaptation of policies to perform new tasks \cite{finn2017model, duan2017one, kirsch2019improving, song2020rapidly}. Our work is related to these approaches, as we leverage a single demonstrations to adapt visual policies; however, our method does not make weight updates through finetuning. Instead, we focus on the Transformer architecture as a means to effectively condition on the demonstration data for adaptation in an in-context manner.

\textbf{One-Shot Imitation Learning with Transformer:}
The Transformer architecture~\cite{vaswani2017attention,lin2022survey} has been highly successful in the field of natural language processing and has been adapted for various other domains, including reinforcement learning and robotic control. Originating in the realm of NLP, the prompt-based approach~\cite{liu2021pre,wei2022chain} has gained traction as an effective method for in-context task adaptation. In this approach, a task-specific prompt is added as a prefix to the input sequence, guiding the model toward the desired output. This design has been adopted in the field of robot learning recently, particularly for adaptation to new tasks encountered during testing, and is framed within the context of one-shot imitation learning. For example, \cite{dasari2021transformers} proposes a one-shot imitation learning method that employs the Transformer architecture to learn new tasks based on a single demonstration from an expert. Subsequent work~\cite{mandi2022towards} further improves generalization capabilities by incorporating temporal contrastive learning. In the context of offline RL, \cite{xu2022prompting} extends the recently proposed Decision Transformer~\cite{chen2021decision} to enable few-shot generalization for unseen tasks. 

The majority of current research in this area studies using a single demonstration to perform unseen testing tasks. However, these approaches often overlook the challenge of adapting to different visual domains, an obstacle that remains significant for important applications such as sim-to-real adaptation. Our aim is to extend the current framework of Transformer-based, demonstration-conditioned policy adaptation to include visual domain adaptation.

\section{Method}
% \textbf{Formulation:}
We frame our problem as a single demonstration conditioned domain adaptation problem for visual policies. Formally, consider a Markov decision process (MDP)~\cite{sutton2018reinforcement} $\mathcal{M}$ parameterized by the tuple $<S,A, O, R,P,\gamma>$, where $S$ and $A$ are the state and action spaces. $P: S \times A \times S \rightarrow [0,1]$ is the transition function, $R: S \times A \rightarrow \mathrm{R}$ is the scalar reward, and $\gamma$ is the discount factor. We assume that the desired visual RL agent cannot directly access state space $S$ but only receives higher-dimensional input $o\in O$ (e.g., pixel images) as observations. The source environment $\mathcal{M}_{src}$ and target environment $\mathcal{M}_{tgt}$ differ only in the mapping from state space $S$ to observation space $O$ due to visual appearance shift, while all other elements remain the same. 

We assume access to an $h$-step demonstration trajectory $\tau^* = (o_0^*, a_o^*, ...,o_{h-1}^*, a_{h-1}^*)$ in the target environment. In this work, we aim to design an algorithm that can efficiently leverage such a single demonstration to perform well on $\mathcal{M}_{tgt}$, which is not known a priori.

\textbf{Policy distillation:}
To concentrate on the issue of domain adaptation through a single demonstration, we divide policy learning and policy adaptation under the framework of policy distillation. In the initial stage, we train a high-performing teacher policy $\pi_t(s): S \rightarrow A$ using a standard RL algorithm directly from state space. The objective is to maximize cumulative episodic reward in the source environment:
\begin{equation}
J(\pi_t) = \mathbb{E}_{\tau \sim \pi_t}[R(\tau)] = \mathbb{E}_{\tau \sim \pi_t}[\sum_{t=0}^{T} \gamma^t r_t],
\end{equation}
where $\tau$ is the rollout trajectory induced by executing teaching policy $\pi_t$. For this stage, we choose Soft Actor-Critic (SAC)~\cite{haarnoja2018soft} as it is a state-of-the-art RL algorithm that is widely used in continuous control tasks.

In the second stage, we train a student policy $\pi_s(o;\tau^*)$ to imitate the optimal actions taken by the teacher, given the observation and a single demonstration $\tau^*$, both of which are under the same domain randomization function $f_{DR}$ from a domain randomization function set $F_{DR}$. This design brings demonstrations and per step observation to the same randomized visual domain and encourages the student policy to learn to adapt its policy based on the demonstration $\tau^*$.

\begin{algorithm}[t]
\caption{Domain adapted visual policy via single demonstration}
\label{dagger}
\begin{algorithmic}[1]
\State \textbf{Input:} $\pi_t, \pi_s$: trained teacher policy and randomly initialized student policy, $\mathcal{D}_{dagger}$: empty dagger dataset, $F$: domain randomization function set, $N$: number of steps for each trial and $K$: number of epochs.
\State Roll out $\pi_t$ to collect initial samples in $\mathcal{D}_{dagger}$
\For{$k = 1, 2, \dots, K$}
  \State Sample domain randomization function $f_{DR} \sim F_{DR}$
  \State Roll out $\pi_t$ to get demonstration $\tau'$ under $f_{DR}$

  \For{$n = 1, 2, \dots, N$}
    \State Roll out $\pi_s$ to collect $o', s'$ samples under $f_{DR}$
    \State Add new data $\mathcal{D}_{dagger} \leftarrow \mathcal{D}_{dagger} \cup \{o', s', \tau'\}$ % define $o_s \sim $
    \State \Comment{Collect data}

    \State Sample tuple $(o, s, \tau^*) \sim \mathcal{D}_{dagger}$ 
    \State Update $\pi_s$ to minimize $\norm{\pi_s(o; \tau^*) - \pi_t(s)}^2$
    \State \Comment{Update student policy}

  \EndFor
\EndFor
\end{algorithmic} 
\end{algorithm} 

The student policy is distilled from the teacher using the DAgger (Dataset Aggregation)~\cite{ross2011reduction} imitation procedure. To begin with, we initialize a DAgger dataset $\mathcal{D}_{dagger}$ by rolling out the teacher policy. In each iteration, we select a domain randomization function $f_{DR} \in F_{DR}$ and apply it to the policy rollout process to collect teacher demonstration $\tau'$ and student rollout observations $o', s'$ from the simulator. The process alternates between (1) updating the student's parameters through gradient descent on a supervised regression loss, and (2) adding more samples to $\mathcal{D}_{dagger}$ using the latest student policy. %The complete pseudo-code of the proposed approach is presented as Algorithm \ref{dagger}.

\textbf{Transformer-based prompt-conditioned visual policy:}
We build our model based on the Transformer architecture. We first process the visual input through a deep convolutional neural network (ConvNet) to extract features. Given an input observation $o$, the ConvNet encoder $E$ generates an output representation $E(o) \in \mathbb{R}^{d_e}$, where $d_e$ is the encoder output dimension. We then define two embedding functions, $f_o : \mathbb{R}^{d_e} \rightarrow \mathbb{R}^{d_h}$ and $f_a : \mathbb{R}^{d_a} \rightarrow \mathbb{R}^{d_h}$ to embed the observation features and actions, respectively, to embedding with dimension of $d_h$. The resulting embeddings $f_o(E(o))$ and $f_a(a)$ are then used as input to the Transformer's encoder layer. To provide sequential information in the trajectory, we incorporate the learnable linear positional encodings $P \in \mathbb{R}^{(h+1) \times d_h}$ as in~\cite{chen2021decision}. We add the positional encodings to the input elements before they are processed by the Transformer encoder, allowing the self-attention mechanism in the Transformer to selectively focus on different parts of the demonstration trajectory and the current observation.

\begin{equation}
\begin{aligned}
\mathrm{Input} &= \mathrm{Concat}\Bigl([f_{o}(E(o_0^*)), f_{a}(a_0^*),\\ & \ldots, (f_{o}(E(o_{h-1}^*)), f_{a}(a_{h-1}^*)), f_{o}(E(o))]\Bigr) + P \\
\end{aligned}
\end{equation}

We then predict the next action to take $\hat{a}$ based on the last element of the output from the Transformer.
To train the student policy, we minimize the $L_2$ distance between the predicted action $\hat{a}$ and the teacher policy's output $a^*$, which is obtained by applying the teacher policy to the state $s$. The loss function $L$ can be formulated as:

\begin{equation}
L(\pi_s) = \norm{\hat{a} - a^*}^2 = \norm{\pi_s(o; \tau^*) - \pi_t(s)}^2,
\end{equation}
where $||\cdot||^2$ denotes the squared $L_2$ distance. By minimizing the imitation loss, the student policy learns to mimic the behavior of the teacher policy. This learning objective results in adaptive policy capable of effectively extracting knowledge from given demonstration and adjusting policies accordingly.

\section{Experiments}

\subsection{Settings} 
\begin{figure}[h!] 
 \centering 
  \includegraphics[width=1.5in]{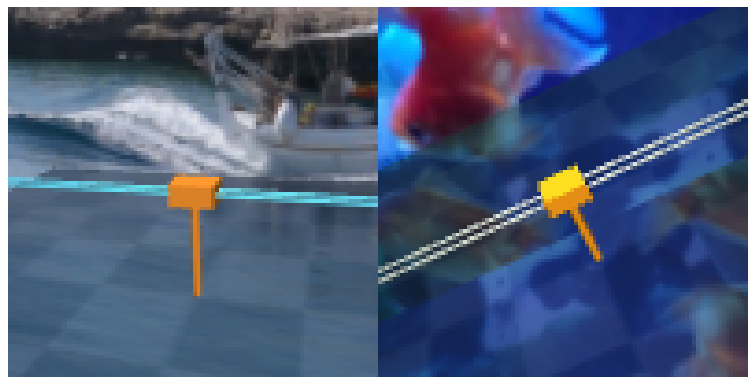}
  \includegraphics[width=1.5in]{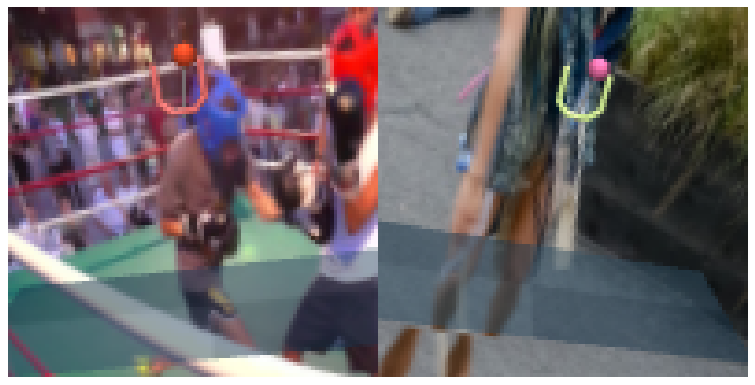}
  \includegraphics[width=1.5in]{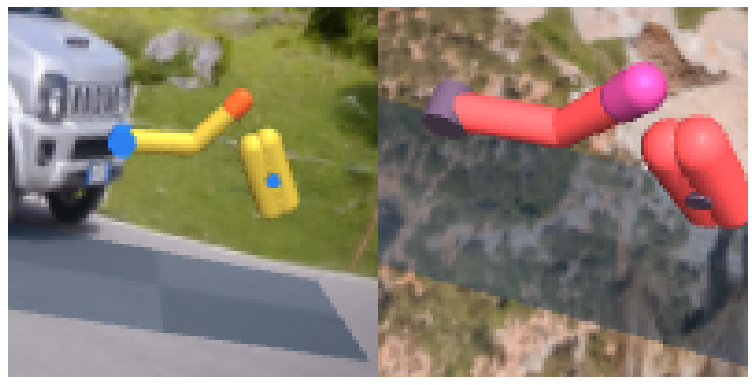}
  \includegraphics[width=1.5in]{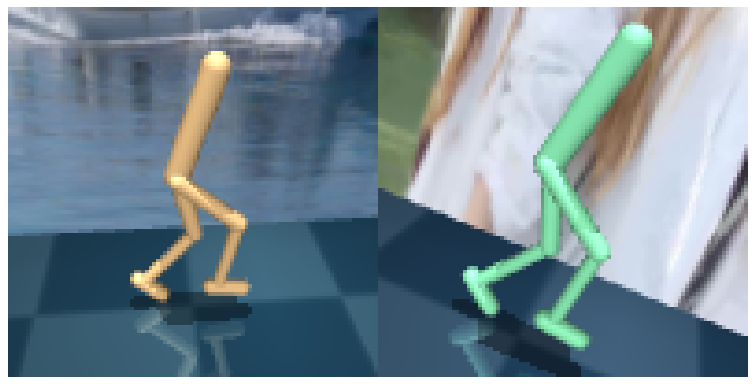}
  \includegraphics[width=1.5in]{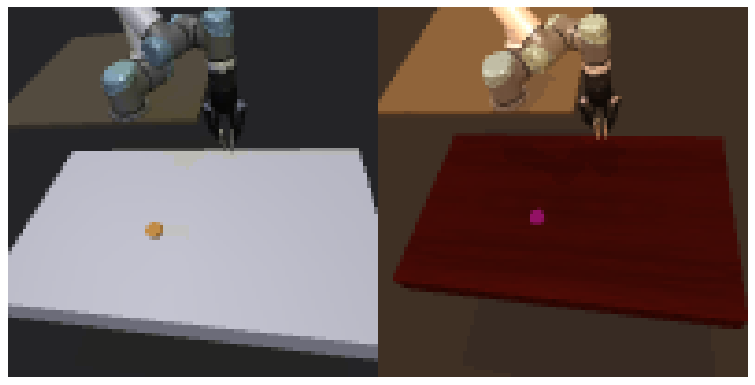}
  \includegraphics[width=1.5in]{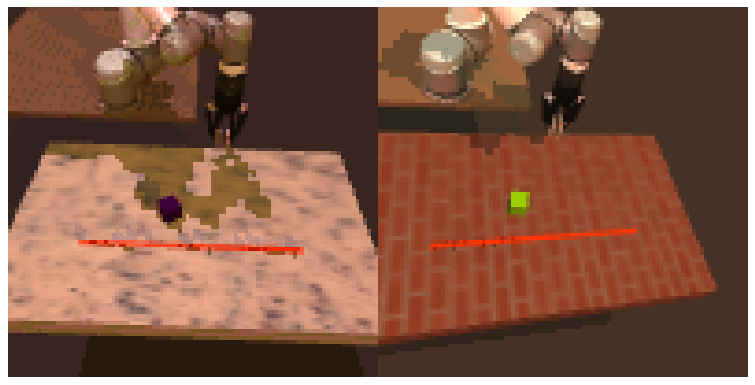}
 \caption{Sample observations in simulated environments. In-distribution image sample is shown left in each pair and out-of-distribution image sample is shown right. First row: \texttt{cartpole swingup}, \texttt{ball\_in\_cup catch}; Second row:  \texttt{finger spin} and  \texttt{walker walk}; Third row: \texttt{UR5 reach} and \texttt{UR5 push}.}
 \label{Distracting CS_sim}
\end{figure}

\textbf{Environments:}
We selected four tasks from the Distracting Control Suite (Distracting CS)~\cite{stone2021distracting} and further implemented two tasks for the UR5 robot in both simulation and real world settings to benchmark our algorithm's performance. Simulation observations can be found in Figure~\ref{Distracting CS_sim}.

Based on the DeepMind Control Suite~\cite{tassa2018deepmind}, Distracting CS is a collection of continuous control tasks designed to challenge an agent's ability to learn in the presence of domain shift. This is introduced by varying levels of changes in colors, backgrounds, and camera poses, making it an ideal benchmark for testing the generalization of reinforcement learning algorithms. We selected four tasks: \texttt{cartpole swingup}, \texttt{ball\_in\_cup catch}, \texttt{finger spin}, and \texttt{walker walk}. The evaluation metric is the episode accumulated reward. We treat the distractions (background, color and camera pose) introduced in this benchmark as domain randomization elements. We train with environments using random distractions of 0.2 intensity as defined in~\cite{tassa2018deepmind}. We then test with random distractions of either 0.2 intensity to evaluate in-distribution (out-of-sample) adaption ability or 0.4 intensity to evaluate out-of-distribution adaption ability.

To further evaluate the applicability of our approach to robotic arm manipulation tasks, we conducted tests with two UR5 robot tasks: reach and push. We have simulated environment implemented with the Mujoco physics engine~\cite{todorov2012mujoco} and a physical UR5 robot for real-world experiments under the sim-to-real setting. The evaluation metric is the task success rate. The success of the reach task is defined as moving the end effector within 5 cm of a colored circular chip, while the success of the push task is defined as pushing a block over a finish line. The policy is trained in simulation with domain randomization including lighting conditions, object colors, table textures, and camera poses. During testing in simulation, we evaluate on a random generated environments with either in-distribution or out-of-distribution domain randomization hyperparameters to test domain adaptation abilities. For testing on physical robot, we also replicate the in-distribution and out-of-distribution settings with corresponding colors and camera poses to evaluate the applicability to real-world scenarios.

\textbf{Baseline Methods:}
In order to benchmark the efficacy of our method in learning domain-adapting visual policies given demonstration, we compare it against three baseline methods: RL with domain randomization (DR)~\cite{tobin2017domain}, policy distillation with domain randomization and policy distillation with model-agnostic meta-learning (MAML)~\cite{finn2017model}. 
% DrQ directly train RL agent on images with domain variations, being the only baseline that does not fall into policy distillation framework. 
DR serves as a baseline that improves robustness solely by expanding the input observation distribution, without utilizing the demonstration at test time. This technique is combined with direct image based RL with SAC~\cite{haarnoja2018soft} and policy distillation respectively. Also used under policy distillation framework, MAML utilizes the demonstration by adapting the policy's weights through gradient descent on observation-action pairs from the demonstration. We use the same policy distillation framework as our method: imitation from trained teacher policy using the DAgger~\cite{ross2011reduction} algorithm.

\begin{table*}[!t]
\centering
\caption{Evaluation in Distracting CS Environments by Episodic Cumulative Rewards (mean $\pm$ standard deviation)}
\label{tab:comparison_Distracting CS}
\begin{tabular}{l|c|c|c|c|c|c|c|c}
% \toprule % top horizontal line
& \multicolumn{4}{c}{In-Distribution Reward} & \multicolumn{4}{c}{Out-of-Distribution Reward} \\
\cmidrule(lr){2-5} \cmidrule(lr){6-9} % partial horizontal line
Task & \multicolumn{1}{c}{RL} & \multicolumn{3}{c}{Policy Distillation} & \multicolumn{1}{c}{RL} & \multicolumn{3}{c}{Policy Distillation} \\
\cmidrule(lr){2-2} \cmidrule(lr){3-5} \cmidrule(lr){6-6} \cmidrule(lr){7-9}
& DR & DR & MAML & PA (ours) & DR & DR & MAML & PA (ours) \\
\midrule % middle horizontal line
\texttt{ball\_in\_cup catch} & $109 \pm 28$ & $718 \pm 70$ & $773 \pm 80$ & $\mathbf{883 \pm 25}$ & $129 \pm 41$ & $372 \pm 61$ & $444 \pm 87$ & $\mathbf{538 \pm 76}$ \\
\texttt{cartpole swingup} & $123 \pm 30$ & $258 \pm 32$ & $234 \pm 75$ & $\mathbf{432 \pm 195}$ & $117 \pm 48$ & $222 \pm 23$ & $203 \pm 43$ & $\mathbf{280 \pm 86}$ \\
\texttt{finger spin} & $13 \pm 6$ & $167 \pm 65$ & $192 \pm 79$ & $\mathbf{408 \pm 210}$ & $12 \pm 4$ & $25 \pm 14$ & $39 \pm 21$ & $\mathbf{103 \pm 43}$ \\
\texttt{walker walk} & $372 \pm 304$ & $304 \pm 101$ & $329 \pm 91$ & $\mathbf{601 \pm 106}$ & $99 \pm 65$ & $145 \pm 32$ & $154 \pm 29$ & $\mathbf{220 \pm 91}$ \\
\bottomrule % bottom horizontal line
\end{tabular}
\end{table*}

\textbf{Algorithm Details:} In our algorithm, PromptAdapt, the teacher policy is learned using SAC~\cite{haarnoja2018soft}, with the state vector serving as input. The actor and critic networks are implemented with a multilayer perceptron (MLP) consisting of two hidden layers, each with a size of 1024 units. The teacher policy is trained with 500K environment steps for Disctracting CS environments and 100K for UR5 environments.
The student policy is constructed using a 3-layer Transformer, accompanied by an 11-layer ConvNet image tokenizer and a linear action tokenizer. To encapsulate temporal information, we stack three consecutive frames for each observation. The student policy is trained with 100K environment steps. During testing in simulation, we use the same teacher policy with groundtruth state access to produce demonstration trajectories; for testing in real-world, we have human tele-operated trajectories as demonstration. 

We implement baseline methods using an 11-layer ConvNet, followed by an MLP with two hidden layers, each of size 1024 units. We ensure that the total number of trainable parameters for our method and both baselines are comparable around 2.7M. Both baseline methods are trained by imitating the same teacher policy by 100K steps as our method. All experiments are repeated across 4 random seeds and evaluated with 20 episodes for simulated experiments and 10 episodes for real-world experiments.

\subsection{Results}

\begin{table*}[!t]
\centering
\caption{Evaluation in simulated UR5 environments by task success rate}
\label{tab:comparison_ur5}
\begin{tabular}{l|c|c|c|c|c|c|c|c}
% \toprule % top horizontal line
& \multicolumn{4}{c}{In-Distribution Reward} & \multicolumn{4}{c}{Out-of-Distribution Reward} \\
\cmidrule(lr){2-5} \cmidrule(lr){6-9} % partial horizontal line
Task & \multicolumn{1}{c}{RL} & \multicolumn{3}{c}{Policy Distillation} & \multicolumn{1}{c}{RL} & \multicolumn{3}{c}{Policy Distillation} \\
\cmidrule(lr){2-2} \cmidrule(lr){3-5} \cmidrule(lr){6-6} \cmidrule(lr){7-9}
& DR & DR & MAML & PA (ours) & DR & DR & MAML & PA (ours) \\
\midrule % middle horizontal line
\texttt{UR5 reach} & $48 \pm 16$ & $90 \pm 12$ & $96 \pm 4$ & $\mathbf{99 \pm 1}$ & $11 \pm 1$ & $52 \pm 10$ & $53 \pm 12$ & $\mathbf{77 \pm 9}$ \\
\texttt{UR5 push} & $39 \pm 44$ & $88 \pm 8$ & $87 \pm 6$ & $\mathbf{96 \pm 1}$ & $9 \pm 14$ & $36 \pm 8$ & $38 \pm 5$ & $\mathbf{68 \pm 9}$ \\
\bottomrule % bottom horizontal line
\end{tabular}
\end{table*}

\textit{ 1. Does the incorporation of a single demonstrations enhance agent performance in both in-distribution and out-of-distribution settings?}

We conduct experiments with Distracting CS benchmarks and simulated UR5 robot tasks to evaluate our method's performance.
As can be seen from the results presented in Table~\ref{tab:comparison_Distracting CS} and Table~\ref{tab:comparison_ur5}, our method consistently outperforms baselines of RL with DR and policy distillation with DR and MAML across all environments, exhibiting superior performance in both in-distribution and out-of-distribution settings. By leveraging the teacher policy embedded in demonstrations, our method is capable of effectively capturing and exploiting the nuances of the visual elements in new environments. The enhanced performance in out-of-distribution settings demonstrates that the agent has strong capability for generalizing to new scenarios.

\textit{ 2. How does the proposed approach improve generalization across different environmental elements?}

\begin{figure}[h!] 
 \centering 
  \subfigure[\texttt{UR5 reach}]{\includegraphics[width=1.5in]{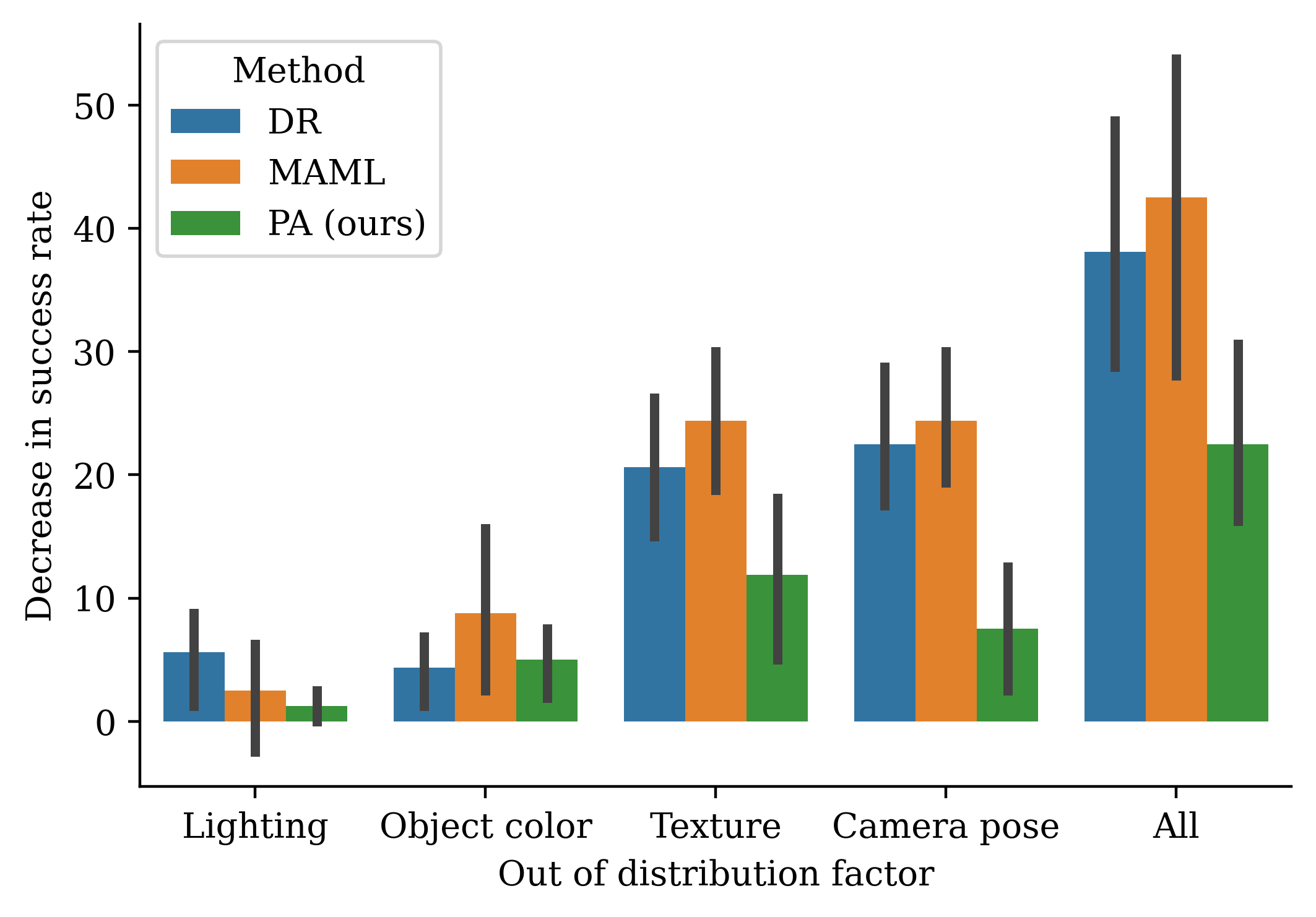}}
  \subfigure[\texttt{UR5 push}]{\includegraphics[width=1.5in]{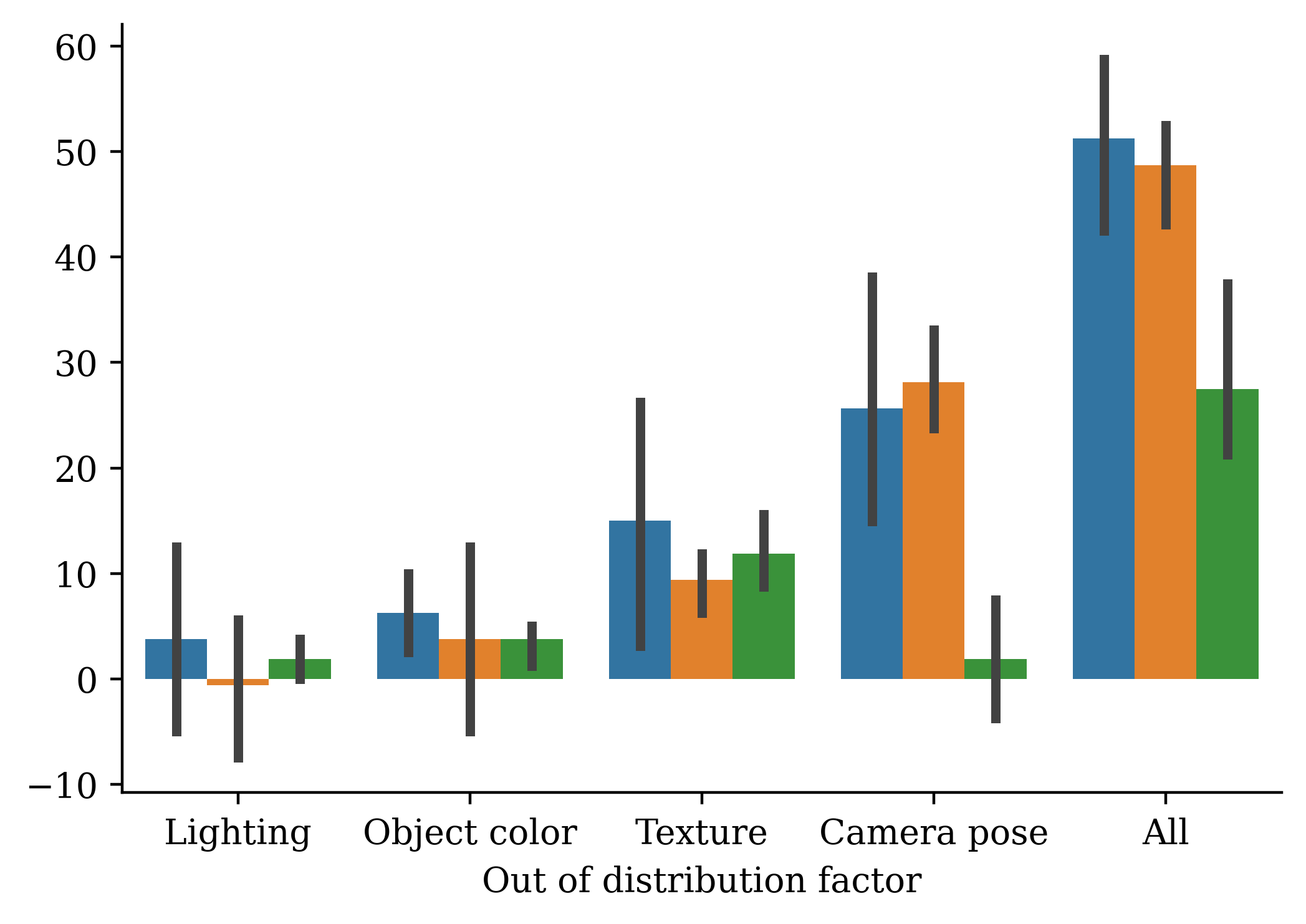}}
 \caption{Ablation study in UR5 simulation environments. Our method leads to lower decrease in success for all out of distribution factors including lighting, object color, table texture and camera pose. Error bar indicates one standard deviation.}
 \label{ur5_generalization_factors}
\end{figure}

We further examine the effectiveness of our method, PromptAdapt (PA), compared to the baselines in terms of generalization over various out-of-distribution factors in the UR5 simulated environments. As illustrated in Figure~\ref{ur5_generalization_factors}, our method consistently outperforms baseline methods across multiple out-of-distribution factors. A significant improvement is observed in the camera pose factor, which is presumably the most challenging aspect in the domain difference. The superior performance of PromptAdapt underscores its ability to capture and adapt to complex changes across domains. We leave RL with DR out of this comparison because it does not perform well in in-distribution setting already due to the difficulty in training RL directly under heavy domain changes.

PromptAdapt also demonstrates improved performance in other out-of-distribution factors, such as lighting and texture. These improvements further support that PromptAdapt can adapt to novel situations with different visual conditions. The consistent performance gains across all out-of-distribution factors suggest that PromptAdapt is a promising approach for creating adaptable agents capable of handling complex tasks in diverse environments.

\begin{figure}[h] 
 \centering 
  \subfigure[In-distribution]{\includegraphics[width=1.5in]{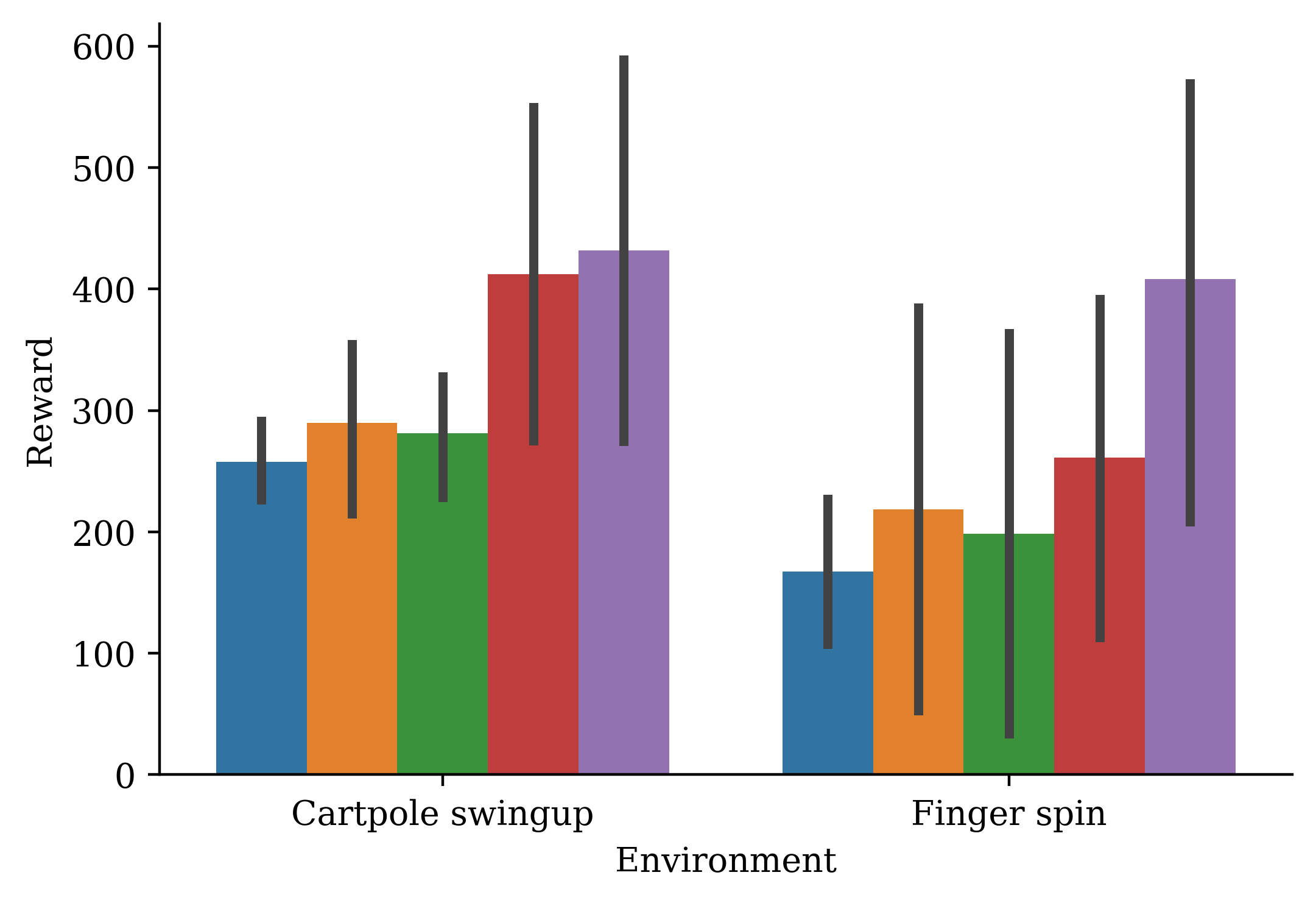}}
  \subfigure[Out-of-distribution]{\includegraphics[width=1.5in]{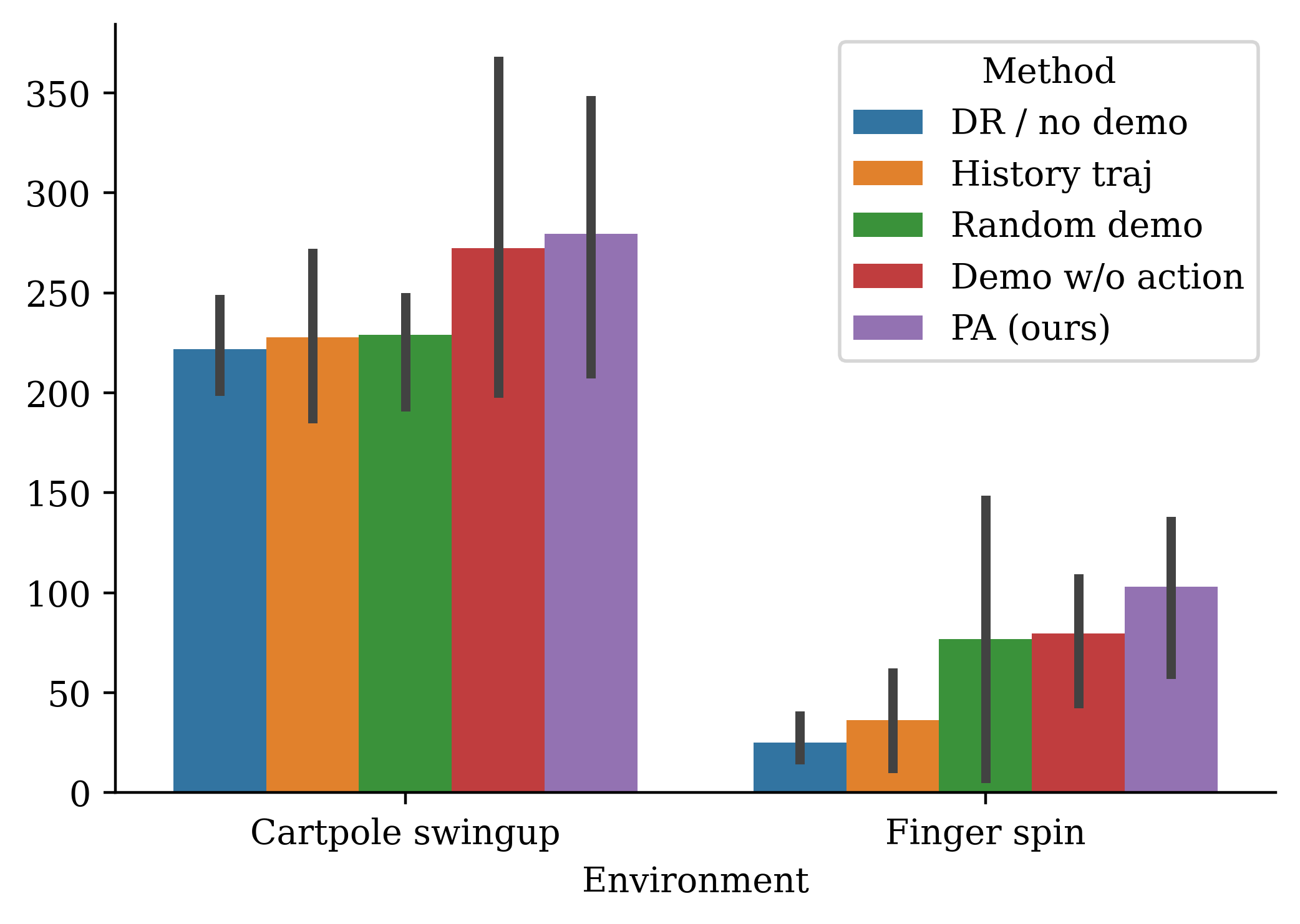}}
 \caption{Ablation study in Distracting CS over different variants in demonstration.} 
 \label{DCS_traj_components}
\end{figure}

\begin{figure*}[h] 
 \centering 
  \includegraphics[width=6in]{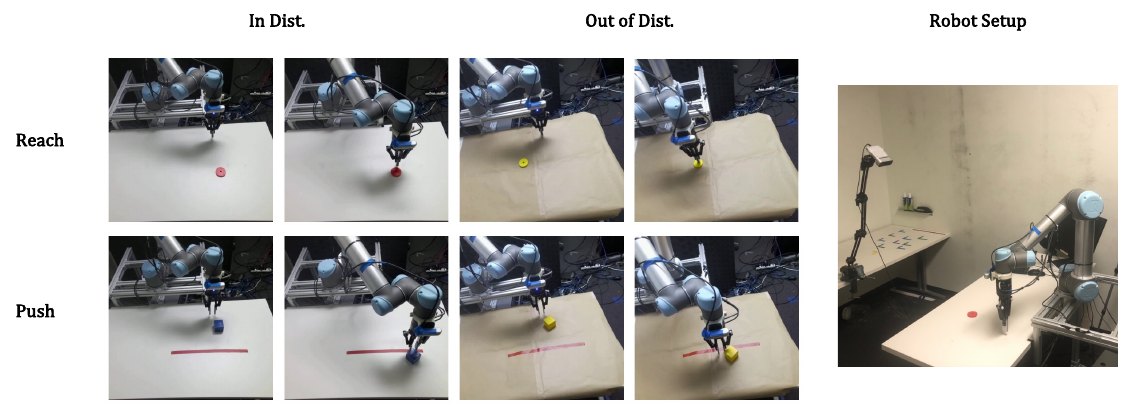}
 \caption{\textbf{Left:} Real-world observations for \texttt{UR5 reach} and \texttt{UR5 push}. \textbf{Right:} Experiment setup.}
 \label{real_world_rollout}
\end{figure*}

\textit{ 3. How do different components in demonstration trajectories contribute to the observed performance of the proposed approach?}

To understand how various components in the trajectory prompt influence the performance of PromptAdapt, we conducted an ablation study in two Distracting CS environments, focusing on the following variants. \texttt{History Trajectory:} replace the demonstration with the previous $h$ steps of student rollout; \texttt{Random Demonstration:} use a random rollout instead of expert demonstration for prompt; \texttt{Demonstration without Action:} use the expert demonstration but excludes action. All variants are trained and tested both in the given configuration.

The findings from our study, as shown in Figure~\ref{DCS_traj_components}, suggest that each of the variants results in some degree of performance drop in both in-distribution and out-of-distribution settings. This implies that a single demonstration contains valuable information about the target environment that isn't fully captured by the history trajectory, a random demonstration, or a demonstration without action. The \texttt{History Trajectory} variant, which utilizes past student rollouts, shows a substantial decrease in performance, suggesting that these rollouts may not fully encapsulate the specifics of the target task. The \texttt{Random Demonstration} variant also shows a similar performance dip, hinting that the quality of the demonstration, rather than merely the presence of test domain trajectories, is critical in effective adaptation. Finally, when action information is excluded in the \texttt{Demonstration without Action} variant, we observe a slight drop in performance, indicating that action sequences in the single demonstration also help adaptation by providing information about transition dynamics and optimal policy. Overall, these results indicate the value of expert demonstrations, inclusive of action sequences, in the successful adaptation of our method.

\textit{ 4. How does our approach apply to real-world?}

We evaluate the real-world applicability of our method under the sim-to-real setting for the two UR5 robot manipulations tasks compared to policy distillation based baseline methods. Demonstration in real-world is recorded by a human tele-operator. 
As demonstrated in Table \ref{tab:comparison_generalization_real_world}, our method consistently outperforms the baseline approaches in the \texttt{UR5 reach} and \texttt{UR5 push}. This highlights the effectiveness of our method in handling both in-distribution and out-of-distribution settings, leading to improved performance under the domain shift from simulation to real-world. Figure~\ref{real_world_rollout} presents visual examples of the real-world experiment.

\begin{table}[ht]
\centering
\caption{Evaluation on real-world robot by success rates.}
\label{tab:comparison_generalization_real_world}
\begin{tabular}{lccc}
&\multicolumn{3}{c}{In distribution success rate} \\ %\hline
\multicolumn{1}{c}{} & \multicolumn{1}{c}{DR} & \multicolumn{1}{c}{MAML} & \multicolumn{1}{c}{\textbf{PA (ours)}} \\ \hline
\texttt{UR5 reach} & 48$\pm$10 & 50$\pm$8 & \textbf{80$\pm$8} \\
\texttt{UR5 push} & 30$\pm$8 & 33$\pm$10 & \textbf{53$\pm$13} \\ \hline
\\
&\multicolumn{3}{c}{Out of distribution success rate} \\ %\hline
\multicolumn{1}{c}{} & \multicolumn{1}{c}{DR} & \multicolumn{1}{c}{MAML} & \multicolumn{1}{c}{\textbf{PA (ours)}} \\ \hline
\texttt{UR5 reach} & 25$\pm$6 & 28$\pm$5 & \textbf{50$\pm$8} \\
\texttt{UR5 push} & 13$\pm$5 & 13$\pm$10 & \textbf{33$\pm$5} \\ \hline
\end{tabular}
\end{table}

\section{Limitation}
  Through our sim-to-real experiments, we have demonstrated that demonstrations provided by human operators during testing can align well with those learned via RL during training. Meanwhile, the proposed method relies on the availability of a demonstration trajectory on the same robot to bridge the domain gap between training and testing environments. This could limit the applicability when only cross-embodiment demonstrations are available (e.g., from a different type of robot). Further analysis and solution to this problem would be in the scope of future work.

\section{Conclusion}

In this paper, we introduced a novel approach to learn domain-adapting visual policies. Our method, utilizing the strengths of policy distillation and the Transformer architecture, efficiently adapts visuomotor policies to target environments using a single demonstration. Through comprehensive experiments, we demonstrate the effectiveness of our method across various simulated and real-world robotics tasks. Ablation studies further illustrate the vital roles of different components in the trajectory prompt, underscoring the significant role of demonstrations in our approach. 
Our proposed method effectively improve the adaptation of policies to novel environments, paving the way for more efficient and effective domain-adapting visuomotor policies.

\newpage
\addtolength{\textheight}{-2.5cm}  
\bibliographystyle{IEEEtran}
\bibliography{IEEEabrv,references}
\addtolength{\textheight}{-6cm}  

\end{document}